\pdfoutput=1
\documentclass[11pt]{article}

\usepackage{acl}

\usepackage{times}
\usepackage{latexsym}

\usepackage[T1]{fontenc}
\usepackage[utf8]{inputenc}

\usepackage{microtype}

\usepackage{inconsolata}

\usepackage{amsmath,amsfonts,amstext}
\usepackage{tikz}
\usetikzlibrary{positioning, calc, decorations.pathreplacing}
\usepackage{booktabs}
\usepackage{multirow}
\usepackage{standalone}
\usepackage{subcaption}
\usepackage{ezutils}

\newcommand{\ours}{DSpERT}
\newcommand{\avestd}[2]{#1\textsubscript{\textpm #2}}
\newcommand{\uparr}{\textcolor{casgreen}{\textuparrow}}
\newcommand{\dwarr}{\textcolor{casred}{\textdownarrow}}

\title{Deep Span Representations for Named Entity Recognition}

\author{Enwei Zhu \and Yiyang Liu \and Jinpeng Li\textsuperscript{\footnotemark[1]}  \\
        Ningbo No.2 Hospital \\
        Ningbo Institute of Life and Health Industry, University of Chinese Academy of Sciences \\ 
    	\texttt{\{zhuenwei,liuyiyang,lijinpeng\}@ucas.ac.cn}}
\begin{document}
\maketitle
{\renewcommand{\thefootnote}{\fnsymbol{footnote}}
\footnotetext[1]{Corresponding author.}
}

\begin{abstract}
Span-based models are one of the most straightforward methods for named entity recognition (NER). Existing span-based NER systems shallowly aggregate the token representations to span representations. However, this typically results in significant ineffectiveness for long entities, a coupling between the representations of overlapping spans, and ultimately a performance degradation. In this study, we propose DSpERT (\textbf{D}eep \textbf{Sp}an \textbf{E}ncoder \textbf{R}epresentations from \textbf{T}ransformers), which comprises a standard Transformer and a span Transformer. The latter uses low-layered span representations as queries, and aggregates the token representations as keys and values, layer by layer from bottom to top. Thus, \ours\ produces span representations of deep semantics. 

With weight initialization from pretrained language models, \ours\ achieves performance higher than or competitive with recent state-of-the-art systems on six NER benchmarks.\footnote{Our code is available at \url{https://github.com/syuoni/eznlp}.} Experimental results verify the importance of the depth for span representations, and show that \ours\ performs particularly well on long-span entities and nested structures. Further, the deep span representations are well structured and easily separable in the feature space. 
\end{abstract}

\section{Introduction}
As a fundamental information extraction task, named entity recognition (NER) requires predicting a set of entities from a piece of text. Thus, the model has to distinguish the entity spans (i.e., positive examples) from the non-entity spans (i.e., negative examples). In this view, it is natural to enumerate all possible spans and classify them into the entity categories (including an extra non-entity category). This is exactly the core idea of span-based approaches~\citep{sohrab-miwa-2018-deep, eberts2020span, yu-etal-2020-named}. 

Analogously to how representation learning matters to image classification~\citep{katiyar-cardie-2018-nested, bengio2013representation, chen2020simple}, it should be crucial to construct good span representations for span-based NER. However, existing models typically build span representations by shallowly aggregating the top/last token representations, e.g., pooling over the sequence dimension~\citep{sohrab-miwa-2018-deep, eberts2020span, shen-etal-2021-locate}, or integrating the starting and ending tokens~\citep{yu-etal-2020-named, li2020empirical}. In that case, the token representations have not been fully interacted before they are fed into the classifier, which impairs the capability of capturing the information of long spans. If the spans overlap, the resulting span representations are technically coupled because of the shared tokens. This causes the representations less distinguishable from the ones of overlapping spans in nested structures. 

Inspired by (probably) the most sophisticated implementation of attention mechanism --- Transformer and BERT~\citep{vaswani2017attention, devlin-etal-2019-bert}, we propose DSpERT, which stands for \textbf{D}eep \textbf{Sp}an \textbf{E}ncoder \textbf{R}epresentations from \textbf{T}ransformers. It consists of a standard Transformer and a \emph{span Transformer}; the latter uses low-layered span representations as queries, and token representations within the corresponding span as keys and values, and thus aggregates token representations layer by layer from bottom to top. Such multi-layered Transformer-style aggregation promisingly produces \emph{deep span representations} of rich semantics, analogously to how BERT yields highly contextualized token representations. 

With weight initialization from pretrained language models (PLMs), \ours\ performs comparably to recent state-of-the-art (SOTA) NER systems on six well-known benchmarks. Experimental results clearly verify the importance of the depth for the span representations. In addition, \ours\ achieves particularly amplified performance improvements against its shallow counterparts\footnote{In this paper, unless otherwise specified, we use ``shallow'' to refer to models that construct span representations by shallowly aggregating (typically top) token representations, although the token representations could be ``deep''.} on long-span entities and nested structures. 

Different from most related work which focuses on the decoder designs~\citep{yu-etal-2020-named, li-etal-2020-unified, shen-etal-2021-locate, li2022unified}, we make an effort to optimize the span representations, but employ a simple and standard neural classifier for decoding. This exposes the \emph{pre-logit representations} that directly determine the entity prediction results, and thus allows further representation analysis widely employed in a broader machine learning community~\citep{van2008visualizing, krizhevsky2012imagenet}. This sheds light on neural NER systems towards higher robustness and interpretability~\citep{ouchi-etal-2020-instance}.

\section{Related Work}
The NER research had been long-term focused on recognizing flat entities. After the introduction of linear-chain conditional random field~\citep{collobert2011natural}, neural sequence tagging models became the \emph{de facto} standard solution for flat NER tasks~\citep{huang2015bidirectional, lample-etal-2016-neural, ma-hovy-2016-end, chiu-nichols-2016-named, zhang-yang-2018-chinese}. 

Recent studies pay much more attention to nested NER, which a plain sequence tagging model struggles with~\citep{ju-etal-2018-neural}. This stimulates a number of novel NER system designs beyond the sequence tagging framework. Hypergraph-based methods extend sequence tagging by allowing multiple tags for each token and multiple tag transitions between adjacent tokens, which is compatible with nested structures~\citep{lu-roth-2015-joint, katiyar-cardie-2018-nested}. Span-based models enumerate candidate spans and classify them into entity categories~\citep{sohrab-miwa-2018-deep, eberts2020span, yu-etal-2020-named}. \citet{li-etal-2020-unified} reformulates nested NER as a reading comprehension task. \citet{shen-etal-2021-locate, shen-etal-2022-parallel} borrow the methods from image object detection to solve nested NER. \citet{yan-etal-2021-unified-generative} propose a generative approach, which encodes the ground-truth entity set as a sequence, and thus reformulates NER as a sequence-to-sequence task. \citet{li2022unified} describe the entity set by word-word relation, and solve nested NER by word-word relation classification. 

The span-based models are probably the most straightforward among these approaches. However, existing span-based models typically build span representations by shallowly aggregating the top token representations from a standard text encoder. Here, the shallow aggregation could be pooling over the sequence dimension~\citep{eberts2020span, shen-etal-2021-locate}, integrating the starting and ending token representations~\citep{yu-etal-2020-named, li2020empirical}, or a concatenation of these results~\citep{sohrab-miwa-2018-deep}. Apparently, shallow aggregation may be too simple to capture the information embedded in long spans; and if the spans overlap, the resulting span representations are technically coupled because of the shared tokens. These ultimately lead to a performance degradation. 

Our \ours\ addresses this issue by multi-layered and bottom-to-top construction of span representations. Empirical results show that such deep span representations outperform the shallow counterpart qualitatively and quantitatively.

\section{Methods}
\paragraph{Deep Token Representations.} Given a $T$-length sequence passed into an $L$-layered $d$-dimensional Transformer encoder~\citep{vaswani2017attention}, the initial token embeddings, together with the potential positional and segmentation embeddings~\citep[e.g., BERT;][]{devlin-etal-2019-bert}, are denoted as $\ma{H}^0 \in \mathbb{R}^{T \times d}$. Thus, the $l$-th ($l = 1, 2, \dots, L$) token representations are: 
\begin{equation}
\ma{H}^{l} = \operatorname{TrBlock} (\ma{H}^{l-1}, \ma{H}^{l-1}, \ma{H}^{l-1}), 
\end{equation}

where $\operatorname{TrBlock} (\ma{Q}, \ma{K}, \ma{V})$ is a Transformer encoder block that takes $\ma{Q} \in \mathbb{R}^{T \times d}$, $\ma{K} \in \mathbb{R}^{T \times d}$, $\ma{V} \in \mathbb{R}^{T \times d}$ as the query, key, value inputs, respectively. It consists of a multi-head attention module and a position-wise feed-forward network (FFN), both followed by a residual connection and a layer normalization. Passing the same matrix, i.e., $\ma{H}^{l-1}$, for queries, keys and values exactly results in self-attention~\citep{vaswani2017attention}. 

The resulting top representations $\ma{H}^L$, computed through $L$ Transformer blocks, are believed to embrace deep, rich and contextualized semantics that are useful for a wide range of tasks. Hence, in a typical neural NLP modeling paradigm, only the top representations $\ma{H}^L$ are used for loss calculation and decoding~\citep{devlin-etal-2019-bert,eberts2020span,yu-etal-2020-named}. 

\begin{figure*}[ht]
    \centering
    \begin{tikzpicture}[recnode/.style = {rectangle, rounded corners=2pt, font=\sffamily\scriptsize}, 
                    cirnode/.style = {circle, thick},
                    marknode/.style ={font=\sffamily\tiny},
                    red7020/.style    = {draw=casred!70, fill=casred!20},
                    yellow7020/.style = {draw=casyellow!70, fill=casyellow!20}, 
                    green7020/.style  = {draw=casgreen!70, fill=casgreen!20}, 
                    blue7020/.style   = {draw=casblue!70, fill=casblue!20}, 
                    brown7020/.style = {draw=C5!70, fill=C5!20}, 
                    gray7020/.style   = {draw=gray!70, fill=gray!20}]
    % Left part
    \node[recnode, thick, yellow7020, minimum width=110pt, minimum height=25pt] at (0pt, 70pt) {Transformer Block}; 
    \node[recnode, thick, red7020, minimum width=110pt, minimum height=15pt] at (0pt, -37.5pt) {Embedding Layer}; 
    \foreach \x in {-32, -16, 0}{
        \node[recnode, thick, red7020, minimum size=10pt] at (\x pt, -60pt) {};
        \draw[-stealth, thick, draw=casred!70] (\x pt, -55pt) -- (\x pt, -45pt);
        \draw[-stealth, thick, draw=casred!70] (\x pt, -30pt) -- (\x pt, -21pt);
        \node[recnode, red7020, minimum width=12pt, minimum height=42pt] at (\x pt, 0pt) {};
        \foreach \y in {-15, -5, 5, 15}{
            \node[cirnode, red7020, minimum size=5pt] at (\x pt, \y pt) {};
        }
        \draw[thick, draw=casred!70, rounded corners=2pt] (\x pt, 21pt) -- ++(0pt, 9pt) -- (0pt, 30pt);
        
        \draw[-stealth, thick, draw=casyellow!70] (\x pt, 82.5pt) -- (\x pt, 99pt);
        \node[recnode, yellow7020, minimum width=12pt, minimum height=42pt] at (\x pt, 120pt) {};
        \foreach \y in {-15, -5, 5, 15}{
            \node[cirnode, yellow7020, minimum size=5pt] at (\x pt, 120+\y pt) {};
        }
        \draw[-stealth, thick, draw=casyellow!70, rounded corners=2pt] (\x pt, 141pt) -- ++(0pt, 6.5pt) -- ++(50-\x/2 pt, 10pt) -- ++(0pt, 10pt);
    }
    \foreach \x in {-48, 16, 32, 48}{
        \node[recnode, thick, red7020, densely dotted, minimum size=10pt] at (\x pt, -60pt) {};
        \draw[-stealth, thick, draw=casred!70, densely dotted] (\x pt, -55pt) -- (\x pt, -45pt);
        \draw[-stealth, thick, draw=casred!70, densely dotted] (\x pt, -30pt) -- (\x pt, -21pt);
        \node[recnode, red7020, densely dotted, minimum width=12pt, minimum height=42pt] at (\x pt, 0pt) {};
        \foreach \y in {-15, -5, 5, 15}{
            \node[cirnode, red7020, densely dotted, minimum size=5pt] at (\x pt, \y pt) {};
        }
        \draw[thick, draw=casred!70, densely dotted, rounded corners=2pt] (\x pt, 21pt) -- ++(0pt, 9pt) -- (0pt, 30pt);
        
        \draw[-stealth, thick, draw=casyellow!70, densely dotted] (\x pt, 82.5pt) -- (\x pt, 99pt);
        \node[recnode, yellow7020, densely dotted, minimum width=12pt, minimum height=42pt] at (\x pt, 120pt) {};
        \foreach \y in {-15, -5, 5, 15}{
            \node[cirnode, yellow7020, densely dotted, minimum size=5pt] at (\x pt, 120+\y pt) {};
        }
        \draw[-stealth, thick, draw=casyellow!70, densely dotted, rounded corners=2pt] (\x pt, 141pt) -- ++(0pt, 6.5pt) -- ++(50-\x/2 pt, 10pt) -- ++(0pt, 10pt);
    }
    \draw[thick, draw=casred!70] (0pt, 30pt) -- (0pt, 47.5pt);
    \foreach \x/\c in {-20/K, 0/V, 20/Q}{
        \draw[-stealth, thick, draw=casred!70, rounded corners=4pt] (0pt, 47.5pt) -- ++(\x pt, 0pt) -- ++(0pt, 10pt);
        \node[marknode] at (\x+6 pt, 52pt) {\c};
    }
    \foreach \x/\c in {130/K, 150/V}{
        \draw[-stealth, thick, draw=casred!70, rounded corners=4pt] (0pt, 42.5pt) -- ++(\x pt, 0pt) -- ++(0pt, 15pt);
        \node[marknode] at (\x+6 pt, 52pt) {\c};
    }
    \foreach \x in {-16}{
        \draw[-stealth, thick, draw=casred!70, rounded corners=4pt] (0pt, 35pt) -- (80 pt, 35pt) -- (80pt, -30pt) -- (150+\x pt, -30pt) -- (150+\x pt, -21pt);
    }
    \foreach \x in {-32, 0, 16, 32}{
        \draw[-stealth, thick, draw=casred!70, densely dotted, rounded corners=4pt] (0pt, 35pt) -- (80 pt, 35pt) -- (80pt, -30pt) -- (150+\x pt, -30pt) -- (150+\x pt, -21pt);
    }
    \node[marknode, align=center, text width=80pt] at (90pt, 25pt) {Initial \\ Aggregation};
    
    % Right part
    \tikzset{shift={(150pt, 0pt)}}
    \node[recnode, thick, green7020, minimum width=90pt, minimum height=25pt] at (0pt, 70pt) {Span Transformer Block}; 
    \foreach \x in {-16}{
        \node[recnode, blue7020, minimum width=12pt, minimum height=42pt] at (\x pt, 0pt) {};
        \foreach \y in {-15, -5, 5, 15}{
            \node[cirnode, blue7020, minimum size=5pt] at (\x pt, \y pt) {};
        }
        \draw[thick, draw=casblue!70, rounded corners=2pt] (\x pt, 21pt) -- ++(0pt, 9pt) -- (0pt, 30pt);
        
        \node[recnode, green7020, minimum width=12pt, minimum height=42pt] at (\x pt, 120pt) {};
        \foreach \y in {-15, -5, 5, 15}{
            \node[cirnode, green7020, minimum size=5pt] at (\x pt, 120+\y pt) {};
        }
        \draw[-stealth, thick, draw=casgreen!70] (\x pt, 82.5pt) -- (\x pt, 99pt);
        \draw[-stealth, thick, draw=casgreen!70, rounded corners=2pt] (\x pt, 141pt) -- ++(0pt, 6.5pt) -- ++(-50-\x/2 pt, 10pt) -- ++(0pt, 10pt);
    }
    \foreach \x in {-32, 0, 16, 32}{
        \node[recnode, blue7020, densely dotted, minimum width=12pt, minimum height=42pt] at (\x pt, 0pt) {};
        \foreach \y in {-15, -5, 5, 15}{
            \node[cirnode, blue7020, densely dotted, minimum size=5pt] at (\x pt, \y pt) {};
        }
        \draw[thick, draw=casblue!70, densely dotted, rounded corners=2pt] (\x pt, 21pt) -- ++(0pt, 9pt) -- (0pt, 30pt);
        
        \node[recnode, green7020, densely dotted, minimum width=12pt, minimum height=42pt] at (\x pt, 120pt) {};
        \foreach \y in {-15, -5, 5, 15}{
            \node[cirnode, green7020, densely dotted, minimum size=5pt] at (\x pt, 120+\y pt) {};
        }
        \draw[-stealth, thick, draw=casgreen!70, densely dotted] (\x pt, 82.5pt) -- (\x pt, 99pt);
        \draw[-stealth, thick, draw=casgreen!70, densely dotted, rounded corners=2pt] (\x pt, 141pt) -- ++(0pt, 6.5pt) -- ++(-50-\x/2 pt, 10pt) -- ++(0pt, 10pt);
    }
    \foreach \x/\c in {20/Q}{
        \draw[-stealth, thick, draw=casblue!70, rounded corners=5pt] (0pt, 30pt) -- (0pt, 42.5pt) -- ++(\x pt, 0pt) -- ++(0pt, 15pt);
        \node[marknode] at (\x+6 pt, 52pt) {\c};
    }
    
    % Whole part
    \tikzset{shift={(-150pt, 0pt)}}
    \node[marknode] at (-60pt, 85pt) {Lx};
    \node[marknode] at (95pt, 85pt) {Lx(K-1)x};
    \draw[draw=gray!70, densely dashed, very thick, rounded corners=5pt] (-60pt, -22.5pt) rectangle (60pt, 22.5pt);
    \node[marknode, align=center, text width=80pt] at (-80pt, 0pt) {Token \\ Embeddings};
    \draw[draw=gray!70, densely dashed, very thick, rounded corners=5pt] (-60pt, 97.5pt) rectangle (60pt, 142.5pt);
    \node[marknode, align=center, text width=80pt] at (-80pt, 120pt) {Deep Token \\ Representations};
    \draw[draw=gray!70, densely dashed, very thick, rounded corners=5pt] (106pt, -22.5pt) rectangle (194pt, 22.5pt);
    \node[marknode, align=center, text width=80pt] at (215pt, 0pt) {Initial Span \\ Representations};
    \draw[draw=gray!70, densely dashed, very thick, rounded corners=5pt] (106pt, 97.5pt) rectangle (194pt, 142.5pt);
    \node[marknode, align=center, text width=80pt] at (215pt, 120pt) {Deep Span \\ Representations};
    
    \foreach \x/\p in {-48/0, -32/1, -16/2, 0/3, 16/4, 32/5, 48/6}{
        \node[marknode] at (\x pt, -70pt) {\p};
    }
    \foreach \x/\p in {-32/0--2, -16/1--3, 0/2--4, 16/3--5, 32/4--6}{
        \node[marknode] at (150+\x pt, -45pt) {\p};
    }
    \node[marknode, align=center, text width=80pt] at (-80pt, -60pt) {Tokens};
    \node[marknode, align=center, text width=80pt] at (-80pt, -70pt) {Positions};
    \node[marknode, align=center, text width=80pt] at (215pt, -45pt) {Span \\ Positions};
    
    \draw[-stealth, draw=gray!70, densely dashed, very thick] (55pt, 70pt) -- (105pt, 70pt);
    \node[marknode, align=center, text width=80pt] at (80pt, 60pt) {Weight \\ Initialization \\ (or Sharing)};
    
    \node[recnode, thick, brown7020, minimum width=140pt, minimum height=15pt] at (75pt, 175pt) {Entity Classifier}; 
\end{tikzpicture}
    \caption{Architecture of \ours. It comprises: (\emph{Left}) a standard $L$-layer Transformer encoder (e.g., BERT); and (\emph{Right}) a span Transformer encoder, where the span representations are the query inputs, and token representations (from the Transformer encoder) are the key/value inputs. There are totally $K-1$ span Transformer encoders, where $K$ is the maximum span size; and each has $L$ layers. The figure specifically displays the case of span size 3; the span of positions 1--3 is highlighted, whereas the others are in dotted lines.}
    \label{fig:deep-span}
\end{figure*}
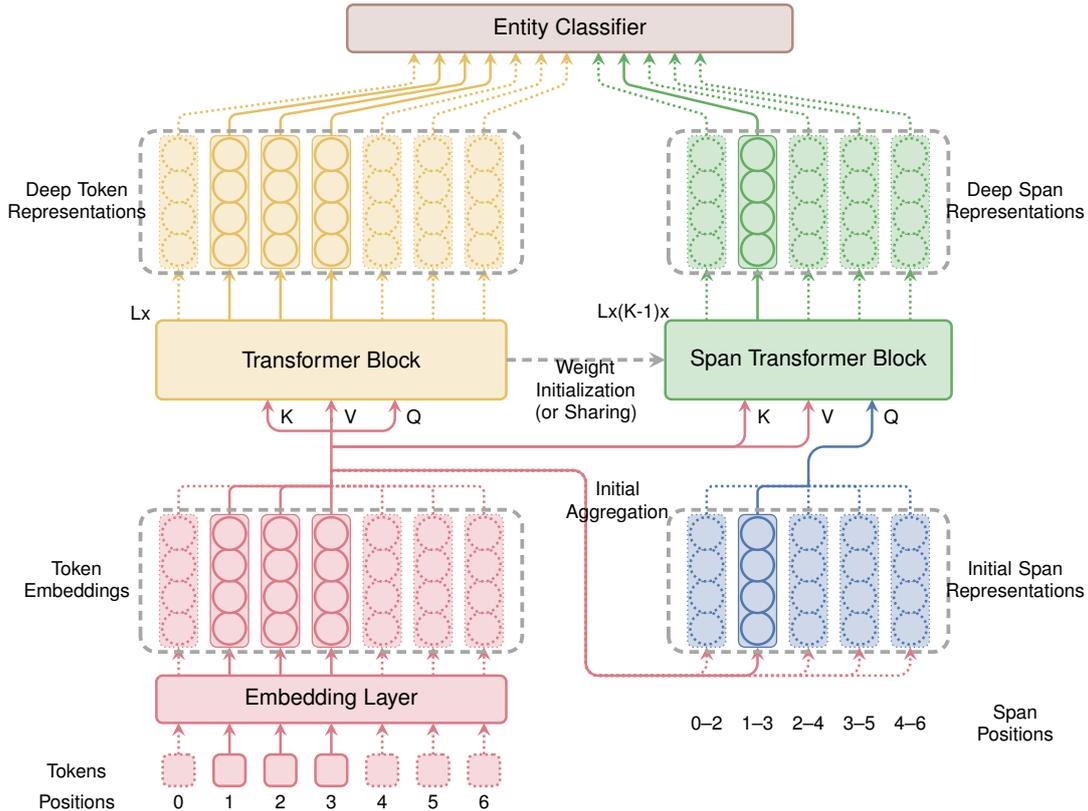

\paragraph{Deep Span Representations.} Figure~\ref{fig:deep-span} presents the architecture of \ours, which consists of a standard Transformer encoder and a \emph{span Transformer encoder}. In a span Transformer of size $k$ ($k = 2, 3, \dots, K$), the initial span representations $\ma{S}^{0,k} \in \mathbb{R}^{(T+k-1) \times d}$ are directly aggregated from the corresponding token embeddings: 
\begin{equation}
\ve{s}^{0,k}_i = \operatorname{Aggregating} (\ma{H}^0_{[i:i+k]}), 
\end{equation}

where $\ve{s}^{0,k}_i \in \mathbb{R}^d$ is the $i$-th vector of $\ma{S}^{0,k}$, and $\ma{H}^0_{[i:i+k]} = [\ve{h}^0_i; \dots; \ve{h}^0_{i+k-1}] \in \mathbb{R}^{k \times d}$ is a slice of $\ma{H}^0$ from position $i$ to position $i+k-1$; $\operatorname{Aggregating} (\cdot)$ is a shallowly aggregating function, such as max-pooling. Check Appendix~\ref{app:aggregating} for more details on alternative aggregating functions used in this study. Technically, $\ve{s}^{0,k}_i$ covers the token embeddings in the span $(i, i+k)$. 

The computation of high-layered span representations imitates that of the standard Transformer. For each span Transformer block, the query is a low-layered span representation vector, and the keys and values are the aforementioned token representation vectors in the positions of that very span. Formally, the $l$-th layer span representations are:
\begin{equation}
\ve{s}^{l,k}_i = \operatorname{SpanTrBlock} (\ve{s}^{l-1,k}_i, \ma{H}^{l-1}_{[i:i+k]}, \ma{H}^{l-1}_{[i:i+k]}), 
\end{equation}

where $\operatorname{SpanTrBlock} (\ma{Q}, \ma{K}, \ma{V})$ shares the exactly same structure with the corresponding Transformer block, but receives different inputs. More specifically, for span $(i, i+k)$, the query is the span representation $\ve{s}^{l-1,k}_i$, and the keys and values are the token representations $\ma{H}^{l-1}_{[i:i+k]}$. Again, the resulting $\ve{s}^{l,k}_i$ technically covers the token representations in the span $(i, i+k)$ on layer $l-1$. 

In our default configuration, the weights of the standard and span Transformers are independent, but initialized from a same PLM. Given the exactly same structure, the weights can be optionally shared between the two modules. This reduces the model parameters, but empirically results in slightly lower performance (See Appendix~\ref{app:ablation}). 

The top span representations $\ma{S}^{L,k}$ are built through $L$ Transformer blocks, which are capable of enriching the representations towards deep semantics. Thus, the representations of overlapping spans are decoupled, and promisingly distinguishable from each other, although they are originally built from $\ma{S}^{0,k}$ --- those shallowly aggregated from token embeddings. This is conceptually analogous to how the BERT uses 12 or more Transformer blocks to produce highly contextualized representations from the original static token embeddings. 

The top span representations are then passed to an entity classifier. Note that we do not construct a unigram span Transformer, but directly borrow the token representations as the span representations of size 1. In other words, 
\begin{equation}
\ma{S}^{L,1} \equiv \ma{H}^{L}. 
\end{equation}

\paragraph{Entity Classifier.} Following \citet{dozat2017deep} and \citet{yu-etal-2020-named}, we introduce a dimension-reducing FFN before feeding the span representations into the decoder. According to the preceding notations, the representation of span $(i, j)$ is $\ve{s}^{L,j-i}_i$, thus, 
\begin{equation}
\ve{z}_{ij} = \operatorname{FFN} (\ve{s}^{L,j-i}_i \oplus \ve{w}_{j-i}), 
\end{equation}
where $\ve{w}_{j-i} \in \mathbb{R}^{d_w}$ is the $(j-i)$-th width embedding from a dedicated learnable matrix; $\oplus$ means the concatenation operation. $\ve{z}_{ij} \in \mathbb{R}^{d_z}$ is the dimension-reduced span representation, which is then fed into a softmax layer: 
\begin{equation} \label{eq:softmax}
\hat{\ve{y}}_{ij} = \softmax (\ma{W} \ve{z}_{ij} + \ve{b}), 
\end{equation}
where $\ma{W} \in \mathbb{R}^{c \times d_z}$ and $\ve{b} \in \mathbb{R}^c$ are learnable parameters, and $\hat{\ve{y}}_{ij} \in \mathbb{R}^c$ is the vector of predicted probabilities over entity types. Note that Eq.~\eqref{eq:softmax} follows the form of a typical neural classification head, which receives a single vector $\ve{z}_{ij}$, and yields the predicted probabilities $\hat{\ve{y}}_{ij}$. Here, the pre-softmax vector $\ma{W} \ve{z}_{ij}$ is called \emph{logits}, and $\ve{z}_{ij}$ is called \emph{pre-logit representation}~\citep{muller2019when}. 

Given the one-hot encoded ground truth $\ve{y}_{ij} \in \mathbb{R}^c$, the model could be trained by optimizing the cross entropy loss for all spans: 
\begin{equation}
\mathcal{L} = -\sum_{0 \leq i < j \leq T} \ve{y}_{ij}^\trans \log(\hat{\ve{y}}_{ij}). 
\end{equation}

We additionally apply the boundary smoothing technique~\citep{zhu-li-2022-boundary}, which is a variant of label smoothing~\citep{szegedy2016rethinking} for span-based NER and brings performance improvements. 

\section{Experiments}
\subsection{Experimental Settings}
\paragraph{Datasets.} We perform experiments on four English nested NER datasets: ACE 2004\footnote{\url{https://catalog.ldc.upenn.edu/LDC2005T09}.}, ACE 2005\footnote{\url{https://catalog.ldc.upenn.edu/LDC2006T06}.}, GENIA~\citep{kim2003genia} and KBP 2017~\citep{ji2017overview}; and two English flat NER datasets: CoNLL 2003~\citep{tjong-kim-sang-de-meulder-2003-introduction} and OntoNotes 5\footnote{\url{https://catalog.ldc.upenn.edu/LDC2013T19}.}. More details on data processing and descriptive statistics are reported in Appendix~\ref{app:datasets}. 

\paragraph{Implementation Details.} To save space, our implementation details are all placed in Appendix~\ref{app:implementation}.

\subsection{Main Results}
Table~\ref{tab:english-nested-res} shows the evaluation results on English nested NER benchmarks. For a fair and reliable comparison to previous SOTA NER systems,\footnote{We exclude previous systems relying on extra training data~\citep[e.g.,][]{li-etal-2020-dice}, external resources~\citep[e.g.,][]{yamada-etal-2020-luke}, extremely large PLMs~\citep[e.g.,][]{yuan-etal-2022-fusing}, or neural architecture search~\citep[e.g.,][]{wang-etal-2021-automated}. } we run \ours\ for five times on each dataset, and report both the best score and the average score with corresponding standard deviation. 

With a \texttt{base}-sized PLM, \ours\ achieves on-par or better results compared with previous SOTA systems. More specifically, the best $F_1$ scores are 88.31\%, 87.42\%, 81.90\% and 87.65\% on ACE 2004, ACE 2005, GENIA and KBP 2017, respectively. Except for ACE 2005, these scores correspond to 0.17\%, 0.13\% and 3.15\% absolute improvements. 

\begin{table}[!t]
    \centering \small
    \begin{tabular}{lccl}
        \toprule
        \multicolumn{4}{c}{ACE 2004} \\
        \midrule
        Model & Prec. & Rec. & ~~F1 \\
        \midrule
        \citet{li-etal-2020-unified}        & 85.05 & 86.32 & 85.98 \\
        \citet{yu-etal-2020-named}          & 87.3~~ & 86.0~~ & 86.7~~ \\
        \citet{yan-etal-2021-unified-generative} & 87.27 & 86.41 & 86.84 \\
        \citet{shen-etal-2021-locate}       & 87.44 & 87.38 & 87.41 \\
        \citet{li2022unified}\textdaggerdbl & 87.33 & 87.71 & 87.52 \\
        \citet{zhu-li-2022-boundary}        & 88.43 & 87.53 & 87.98 \\
        \citet{shen-etal-2022-parallel}     & 88.48 & 87.81 & 88.14 \\
        \midrule
        \ours \textdagger                   & 88.29 & 88.32 & \textbf{88.31} \\
        \ours \textdaggerdbl                & 87.90 & 88.21 & \avestd{88.05}{0.18} \\
        \bottomrule
        \toprule
        \multicolumn{4}{c}{ACE 2005} \\
        \midrule
        Model & Prec. & Rec. & ~~F1 \\
        \midrule
        \citet{li-etal-2020-unified}        & 87.16 & 86.59 & 86.88 \\
        \citet{yu-etal-2020-named}          & 85.2~~ & 85.6~~ & 85.4~~ \\
        \citet{yan-etal-2021-unified-generative} & 83.16 & 86.38 & 84.74 \\
        \citet{shen-etal-2021-locate}       & 86.09 & 87.27 & 86.67 \\
        \citet{li2022unified}\textdaggerdbl & 85.03 & 88.62 & 86.79 \\
        \citet{zhu-li-2022-boundary}        & 86.25 & 88.07 & 87.15 \\
        \citet{shen-etal-2022-parallel}     & 86.27 & 88.60 & \textbf{87.42} \\
        \midrule
        \ours \textdagger                   & 87.01 & 87.84 & \textbf{87.42} \\
        \ours \textdaggerdbl                & 85.73 & 88.19 & \avestd{86.93}{0.49} \\
        \bottomrule
        \toprule
        \multicolumn{4}{c}{GENIA} \\
        \midrule
        Model & Prec. & Rec. & ~~F1 \\
        \midrule
        \citet{yu-etal-2020-named}*         & 81.8~~ & 79.3~~ & 80.5~~ \\
        \citet{yan-etal-2021-unified-generative} & 78.87 & 79.6 & 79.23 \\
        \citet{shen-etal-2021-locate}*      & 80.19 & 80.89 & 80.54 \\
        \citet{li2022unified}\textdaggerdbl & 83.10 & 79.76 & 81.39 \\
        \citet{shen-etal-2022-parallel}*    & 83.24 & 80.35 & 81.77 \\
        \midrule
        \ours \textdagger                   & 82.31 & 81.49 & \textbf{81.90} \\
        \ours \textdaggerdbl                & 81.72 & 81.21 & \avestd{81.46}{0.25} \\
        \bottomrule
        \toprule
        \multicolumn{4}{c}{KBP 2017} \\
        \midrule
        Model & Prec. & Rec. & ~~F1 \\
        \midrule
        \citet{li-etal-2020-unified}        & 82.33 & 77.61 & 80.97 \\
        \citet{shen-etal-2021-locate}       & 85.46 & 82.67 & 84.05 \\
        \citet{shen-etal-2022-parallel}     & 85.67 & 83.37 & 84.50 \\
        \midrule
        \ours \textdagger                   & 87.37 & 87.93 & \textbf{87.65} \\
        \ours \textdaggerdbl                & 87.00 & 87.33 & \avestd{87.16}{0.50} \\
        \bottomrule
    \end{tabular}
    \caption{Results of English nested entity recognition. * means that the model is trained with both the training and development splits. \textdagger~means the best score; \textdaggerdbl~means the average score of multiple independent runs; the subscript number is the corresponding standard deviation.}
    \label{tab:english-nested-res}
\end{table}

Table~\ref{tab:english-flat-res} presents the results on English flat NER datasets. The best $F_1$ scores are 93.70\% and 91.76\% on CoNLL 2003 and OntoNotes 5, respectively. These scores are slightly higher than those reported by previous literature. 

Appendix~\ref{app:categorical} further lists the category-wise $F_1$ scores; the results show that \ours\ can consistently outperform the biaffine model, a classic and strong baseline, across most entity categories. Appendix~\ref{app:chinese-flat-res} provides additional experimental results on Chinese NER, suggesting that the effectiveness of \ours\ is generalizable across languages. 

Overall, \ours\ shows strong and competitive performance on both the nested and flat NER tasks. Given the long-term extensive investigation and experiments on these datasets by the NLP community, the seemingly marginal performance improvements are still notable. %In addition, compared with \citet{li2022unified} who report average evaluation results, \ours\ can achieve on-par or higher average $F_1$ scores on most datasets. This suggests that our model performs robustly. 

\begin{table}[!t]
    \centering \small
    \begin{tabular}{lccl}
        \toprule
        \multicolumn{4}{c}{CoNLL 2003} \\
        \midrule
        Model & Prec. & Rec. & ~~F1 \\
        \midrule
        \citet{peters-etal-2018-deep}\textdaggerdbl & -- & -- & \avestd{92.22}{0.10} \\
        \citet{devlin-etal-2019-bert}       & -- & -- & 92.8~~ \\
        \citet{li-etal-2020-unified}        & 92.33 & 94.61 & 93.04 \\
        \citet{yu-etal-2020-named}*         & 93.7~~ & 93.3~~ & 93.5~~ \\
        \citet{yan-etal-2021-unified-generative}* & 92.61 & 93.87 & 93.24 \\
        \citet{li2022unified}\textdaggerdbl & 92.71 & 93.44 & 93.07 \\
        \citet{zhu-li-2022-boundary}        & 93.61 & 93.68 & 93.65 \\
        \citet{shen-etal-2022-parallel}*    & 93.29 & 92.46 & 92.87 \\
        \midrule
        \ours \textdagger                   & 93.48 & 93.93 & \textbf{93.70} \\
        \ours \textdaggerdbl                & 93.39 & 93.88 & \avestd{93.64}{0.06} \\
        \bottomrule
        \toprule
        \multicolumn{4}{c}{OntoNotes 5} \\
        \midrule
        Model & Prec. & Rec. & ~~F1 \\
        \midrule
        \citet{li-etal-2020-unified}        & 92.98 & 89.95 & 91.11 \\
        \citet{yu-etal-2020-named}          & 91.1~~ & 91.5~~ & 91.3~~ \\
        \citet{yan-etal-2021-unified-generative} & 89.99 & 90.77 & 90.38 \\
        \citet{li2022unified}\textdaggerdbl & 90.03 & 90.97 & 90.50 \\
        \citet{zhu-li-2022-boundary}        & 91.75 & 91.74 & 91.74 \\
        \citet{shen-etal-2022-parallel}     & 91.43 & 90.73 & 90.96 \\
        \midrule
        \ours \textdagger                   & 91.46 & 92.05 & \textbf{91.76} \\
        \ours \textdaggerdbl                & 90.87 & 91.25 & \avestd{91.06}{0.26} \\
        \bottomrule
    \end{tabular}
    \caption{Results of English flat entity recognition. * means that the model is trained with both the training and development splits. \textdagger~means the best score; \textdaggerdbl~means the average score of multiple independent runs; the subscript number is the corresponding standard deviation.}
    \label{tab:english-flat-res}
\end{table}

\subsection{Ablation Studies} \label{subsec:ablation}
We perform ablation studies on three datasets, i.e., ACE 2004, GENIA and CoNLL 2003, covering flat and nested, common and domain-specific corpora. 

\paragraph{Depth of Span Representations.} As previously highlighted, our core argument is that the deep span representations, which are computed throughout the span Transformer blocks, embrace deep and rich semantics and thus outperform the shallow counterparts. %, i.e., those span representations shallowly aggregated from token representations. 

To validate this point, Table~\ref{tab:depth} compares \ours\ to the models with a shallow setting, where the span representations are aggregated from the top token representations by max-pooling, mean-pooling, multiplicative attention or additive attention (See Appendix~\ref{app:aggregating} for details). All the models are trained with the same recipe used in our main experiments. It shows that the 12-layer deep span representations achieve higher performance than its shallow counterparts equipped with any potential aggregating function, across all datasets. 

We further run \ours\ with $\tilde{L}$ ($\tilde{L} < L$) span Transformer blocks, where the initial aggregation happens at the $(L-\tilde{L})$-th layer and the span Transformer corresponds to the top/last $\tilde{L}$ Transformer blocks. These models may be thought of as intermediate configurations between fully deep span representations and fully shallow ones. As displayed in Table~\ref{tab:depth}, the $F_1$ score in general experiences a monotonically increasing trend when depth $\tilde{L}$ increases from 2 to 12; this pattern holds for all three datasets. These results further strengthen our argument that the depth positively contributes to the quality of span representations. 

Appendix~\ref{app:ablation} provides extensive ablation studies evaluating other components. 

\begin{table}[!t]
    \centering \small
    \begin{tabular}{lccc}
        \toprule
        Depth & ACE04 & GENIA & CoNLL03 \\
        \midrule
        \multicolumn{4}{l}{Shallow agg. ($\tilde{L}$ = 0)} \\
        \quad w/ max-pooling     & \avestd{82.22}{0.64} & \avestd{79.44}{0.20} & \avestd{92.99}{0.32} \\
        \quad w/ mean-pooling    & \avestd{80.90}{0.28} & \avestd{73.83}{0.42} & \avestd{91.97}{0.14} \\
        \quad w/ mul. attention  & \avestd{84.38}{0.58} & \avestd{76.54}{2.70} & \avestd{93.21}{0.16} \\
        \quad w/ add. attention  & \avestd{83.73}{0.52} & \avestd{76.23}{3.27} & \avestd{93.05}{0.04} \\
        \midrule
        \ours \\
        \quad $\tilde{L}$ = 2  & \avestd{87.87}{0.13} & \avestd{80.66}{0.36} & \avestd{93.30}{0.09} \\
        \quad $\tilde{L}$ = 4  & \avestd{87.88}{0.41} & \avestd{80.88}{0.39} & \avestd{93.38}{0.08} \\
        \quad $\tilde{L}$ = 6  & \avestd{87.81}{0.13} & \avestd{81.01}{0.22} & \avestd{93.40}{0.13} \\
        \quad $\tilde{L}$ = 8  & \avestd{88.00}{0.22} & \avestd{81.13}{0.25} & \avestd{93.48}{0.08} \\ 
        \quad $\tilde{L}$ = 10 & \avestd{88.00}{0.21} & \avestd{81.12}{0.14} & \avestd{93.51}{0.10} \\
        \quad \underline{$\tilde{L}$ = 12} & \avestd{\textbf{88.05}}{0.18} & \avestd{\textbf{81.46}}{0.25} & \avestd{\textbf{93.64}}{0.06} \\
        \bottomrule
    \end{tabular}
    \caption{The effect of depth. The underlined specification is the one used in our main experiments. All the results are average scores of five independent runs, with subscript standard deviations.}
    \label{tab:depth}
\end{table}

\subsection{Effect on Long-Span Entities}
The recognition of long-span entities is a long-tail and challenging problem. Taking ACE 2004 as an example, the ground-truth entities longer than 10 tokens only account for 2.8\%, and the maximum length reaches 57. Empirical evidence also illustrates that existing NER models show relatively weak performance on long entities~\citep[e.g.,][]{shen-etal-2021-locate,yuan-etal-2022-fusing}. 

\begin{figure*}[t]
    \centering
    \begin{subfigure}{0.32\textwidth}
    \centering
    \includegraphics[width=\textwidth]{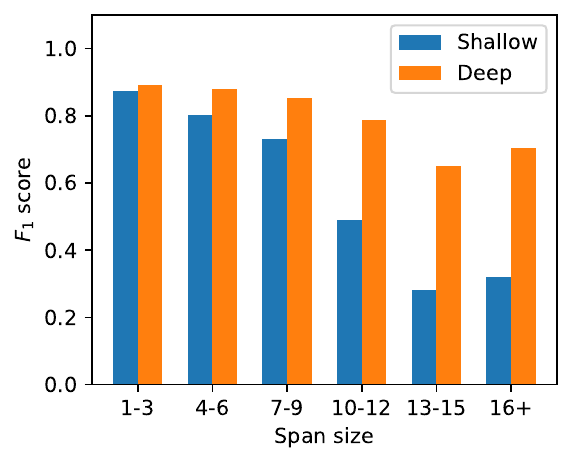}
    \caption{ACE 2004}
    \end{subfigure}
    \begin{subfigure}{0.32\textwidth}
    \centering
    \includegraphics[width=\textwidth]{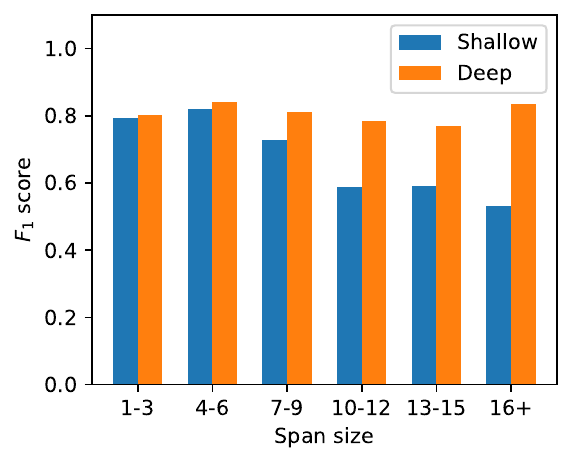}
    \caption{GENIA}
    \end{subfigure}
    \begin{subfigure}{0.32\textwidth}
    \centering
    \includegraphics[width=\textwidth]{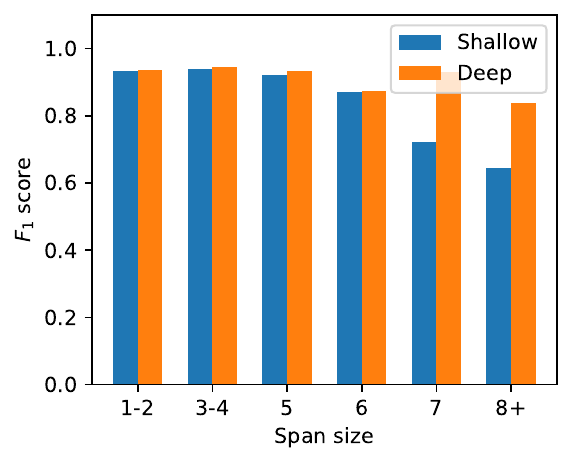}
    \caption{CoNLL 2003}
    \end{subfigure}
    \caption{$F_1$ scores on spans of different lengths. All the results are average scores of five independent runs.}
    \label{fig:f1-by-lens}
\end{figure*}

Figure~\ref{fig:f1-by-lens} presents the $F_1$ scores grouped by different span lengths. In general, the models based on shallow span representations perform relatively well on the short entities, but struggle for the long ones. However, \ours\ show much higher $F_1$ scores on the long entities, without any performance sacrifice on the short ones. For ACE 2004, \ours\ outperforms its shallow counterpart by 2\%--12\% absolute $F_1$ score on spans shorter than 10, while this difference exceeds 30\% for spans longer than 10. Similar patterns are observed on GENIA and CoNLL 2003. 

Conceptually, a longer span contains more information, so it would be more difficult to be encoded into a fixed-length vector, i.e., the span representation. According to our experimental results, the shallow aggregation fails to fully preserve the semantics in the original token representations, especially for long spans. The \ours, however, allows complicated interactions between tokens through multiple layers; in particular, longer spans experience more interactions. This mechanism amplifies the performance gain on long entities.

\subsection{Effect on Nested Structures}
Even in nested NER datasets, the nested entities are less than the flat ones (See Table~\ref{tab:descriptive-statstics}). To deliberately investigate the performance on nested entities, we look into two subsets of spans that are directly related to nested structures: (1) \emph{Nested}: the spans that are nested inside a ground-truth entity which covers other ground-truth entities; (2) \emph{Covering}: the spans that cover a ground-truth entity which is nested inside other ground-truth entities.

For example, in sentence ``Mr. John Smith graduated from New York University last year'', a location entity ``New York'' is nested in an organization entity ``New York University''. A model is regarded to well handle nested structures if it can: (1) distinguish ``New York'' from other negative spans inside the outer entity ``New York University'', i.e., those in the \emph{Nested} subset; and (2) distinguish ``New York University'' from other negative spans covering the inner entity ``New York'', i.e., those in the \emph{Covering} subset. 

\begin{figure*}[t]
\centering
    \begin{subfigure}{0.32\textwidth}
    \centering
    \includegraphics[width=\textwidth]{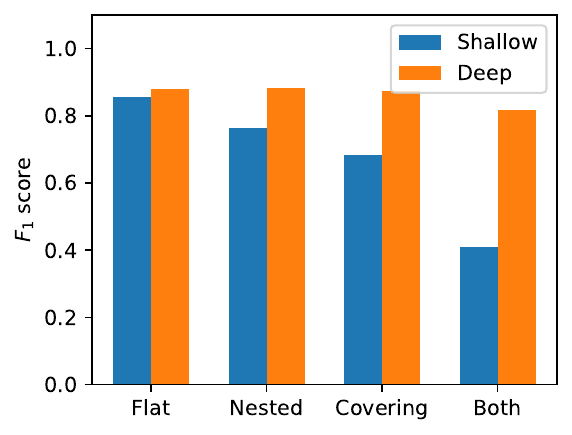}
    \caption{ACE 2004}
    \end{subfigure}
    \begin{subfigure}{0.32\textwidth}
    \centering
    \includegraphics[width=\textwidth]{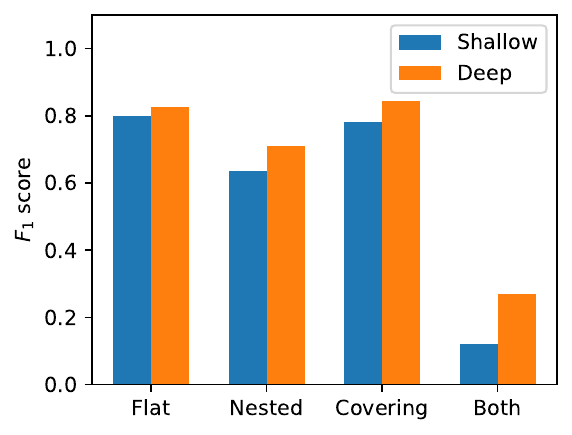}
    \caption{GENIA}
    \end{subfigure}
    \caption{$F_1$ scores on spans with different nested structures. 
    ``Nested'' means the spans that are nested inside a ground-truth entity which covers other ground-truth entities; 
    ``Covering'' means the spans that cover a ground-truth entity which is nested inside other ground-truth entities; 
    ``Both'' means the spans that are both ``Nested'' and ``Covering''; 
    ``Flat'' means the spans that are neither ``Nested'' nor ``Covering''. 
    All the results are average scores of five independent runs.}
    \label{fig:f1-by-nested}
\end{figure*}

Figure~\ref{fig:f1-by-nested} depicts the $F_1$ scores grouped by different nested structures. Consistent to a common expectation, nested structures create significant difficulties for entity recognition. Compared to the flat ones, a shallow span-based model encounters a substantial performance degradation on the nested structures, especially on the spans in both the \emph{Nested} and \emph{Covering} subsets. On the other hand, our deep span representations perform much better. For ACE 2004, \ours\ presents 2\% higher absolute $F_1$ score than the shallow model on flat spans, but achieves about 40\% higher score on spans in both the \emph{Nested} and \emph{Covering} subsets. The experiments on GENIA report similar results, although the difference in performance gain becomes less substantial.  

As previously emphasized, shallowly aggregated span representations are technically coupled if the spans overlap. This explains why such models perform poorly on nested structures. Our model addresses this issue by deep and multi-layered construction of span representations. Implied by the experimental results, deep span representations are less coupled and more easily separable for overlapping spans.

\section{Analysis of Pre-Logit Representations}
For a neural classification model, the logits only relate to the pre-defined classification categories, while the pre-logit representations contain much richer information~\citep{krizhevsky2012imagenet, wu2018unsupervised}. Hence, the analysis of pre-logit representations has become a popular tool in machine learning research~\citep{van2008visualizing, muller2019when, chen2020simple}. 

However, such analysis is incompatible with most neural NER systems except for the span-based ones. Our \ours\ employs a standard classification head, which exposes the pre-logit representations and thus allows the analysis. In this section, we investigate the pre-logit span representations, i.e., $\ve{z}_{ij}$ in Eq.~\eqref{eq:softmax}, providing more insights into why the deep span representations outperform the shallow counterparts. 

\subsection{Decoupling Effect on Overlapping Spans}
The effectiveness of \ours\ on nested structures primarily attributes to its decoupling effect on the representations between overlapping spans. To support this argument, we compare the coupling strengths between span representations with different overlapping levels. 

Specifically, we define the \emph{overlapping ratio} $\alpha$ of two given spans as the proportion of the shared tokens in the spans, and then categorize the span pairs into three scenarios: \emph{non-overlapping} ($\alpha = 0$), \emph{weakly overlapping} ($0 < \alpha \leq 0.5$) and \emph{strongly overlapping} ($0.5 < \alpha < 1$). Table~\ref{tab:decoupling} reports the cosine similarities of the representations between entities and their neighboring spans, categorized by overlapping ratios. In general, \ours\ and the shallow model have comparable similarity values on the non-overlapping spans; and \ours\ show slightly higher values on the overlapping spans. However, the shallow model produces significantly higher similarities for stronger overlapping levels. Hence, shallow models yield coupled representations for overlapping spans, while \ours\ can effectively decouple the representations and thus lead to the performance improvement, in particular on the nested entities. 

\begin{table}[t]
    \centering \small
    \begin{tabular}{lccc}
        \toprule
         & \multicolumn{3}{c}{ACE 2004} \\
        \cmidrule(lr){2-4} 
         & Shallow & Deep & $\Delta$ \\
        \midrule
        Non-overlapping      & \avestd{0.45}{0.05} & \avestd{0.40}{0.04} & \textcolor{casred}{-0.05} \\
        Weakly overlapping   & \avestd{0.65}{0.02} & \avestd{0.41}{0.04} & \textcolor{casred}{-0.24} \\
        Strongly overlapping & \avestd{0.76}{0.01} & \avestd{0.44}{0.04} & \textcolor{casred}{-0.32} \\
        \bottomrule 
        \toprule
         & \multicolumn{3}{c}{GENIA} \\
        \cmidrule(lr){2-4} 
         & Shallow & Deep & $\Delta$ \\
        \midrule
        Non-overlapping      & \avestd{0.39}{0.02} & \avestd{0.35}{0.03} & \textcolor{casred}{-0.04} \\
        Weakly overlapping   & \avestd{0.59}{0.01} & \avestd{0.38}{0.03} & \textcolor{casred}{-0.21} \\
        Strongly overlapping & \avestd{0.72}{0.00} & \avestd{0.42}{0.04} & \textcolor{casred}{-0.30} \\
        \bottomrule 
        \toprule
         & \multicolumn{3}{c}{CoNLL 2003} \\
        \cmidrule(lr){2-4} 
         & Shallow & Deep & $\Delta$ \\
        \midrule
        Non-overlapping      & \avestd{0.31}{0.02} & \avestd{0.40}{0.02} & \textcolor{casgreen}{+0.09} \\
        Weakly overlapping   & \avestd{0.58}{0.01} & \avestd{0.46}{0.02} & \textcolor{casred}{-0.12} \\
        Strongly overlapping & \avestd{0.69}{0.01} & \avestd{0.52}{0.02} & \textcolor{casred}{-0.17} \\
        \bottomrule 
    \end{tabular}
    \caption{Cosine similarities of the representations between entity spans and their neighboring spans. 
    Non-/weakly/strongly overlapping means that the overlapping ratio is 0/0--0.5/0.5--1, respectively. 
    All the metrics are first averaged within each experiment, and then averaged over five independent experiments, reported with subscript standard deviations.}
    \label{tab:decoupling}
\end{table}

\subsection{$\ell_2$-Norm and Cosine Similarity}
We calculate the $\ell_2$-norm and cosine similarity of the span representations. As presented in Table~\ref{tab:l2-and-cos}, the deep span representations have larger $\ell_2$-norm than those of the shallow counterpart. Although the variance of representations inevitably shrinks during the aggregating process from the perspective of statistics, this result implies that deep span representations are less restricted and thus able to flexibly represent rich semantics. In addition, the deep span representations are associated with higher within-class similarities and lower between-class similarities, suggesting that the representations are more tightly clustered within each category, but more separable between categories. Apparently, this nature contributes to the high classification performance. 

We further investigate the pre-logit weight, i.e., $\ma{W}$ in Eq.~\eqref{eq:softmax}. First, the trained \ours\ has a pre-logit weight with a smaller $\ell_2$-norm. According to a common understanding of neural networks, smaller norm implies that the model is simpler and thus more generalizable. 

Second, as indicated by \citet{muller2019when}, a typical neural classification head can be regarded as a \emph{template-matching} mechanism, where each row vector of $\ma{W}$ is a \emph{template}.\footnote{Conceptually, if vector $\ve{z}_{ij}$ is most matched/correlated with the $k$-th row vector of $\ma{W}$, then the $k$-th logit will be most activated. Refer to \citet{muller2019when} for more details.} Under this interpretation, each template ``stands for'' the overall direction of the span representations of the corresponding category in the feature space. As shown in Table~\ref{tab:l2-and-cos}, the absolute cosine values between the templates of \ours\ are fairly small. In other words, the templates are approximately orthogonal, which suggests that different entity categories are uncorrelated and separately occupy distinctive subareas in the feature space. This pattern, however, is not present for the shallow models.

\begin{table*}[!t]
    \centering \small
    \begin{tabular}{lcccccc}
        \toprule
         & \multicolumn{2}{c}{ACE 2004} & \multicolumn{2}{c}{GENIA} & \multicolumn{2}{c}{CoNLL 2003} \\
        \cmidrule(lr){2-3} \cmidrule(lr){4-5} \cmidrule(lr){6-7} 
         & Shallow & Deep & Shallow & Deep & Shallow & Deep \\
        \midrule
        Pre-logit representations \\
        \quad $\ell_2$-norm                  & \avestd{9.58}{0.17} & \avestd{14.55}{0.33}\uparr  & \avestd{9.89}{0.18} & \avestd{11.69}{0.52}\uparr  & \avestd{9.08}{0.17} & \avestd{13.49}{0.50}\uparr  \\
        \quad Cosine within pos class        & \avestd{0.70}{0.02} & \avestd{~~0.80}{0.01}\uparr & \avestd{0.78}{0.01} & \avestd{~~0.83}{0.01}\uparr & \avestd{0.79}{0.01} & \avestd{~~0.88}{0.00}\uparr \\
        \quad Cosine between pos vs. pos     & \avestd{0.45}{0.02} & \avestd{~~0.37}{0.01}\dwarr & \avestd{0.54}{0.02} & \avestd{~~0.29}{0.01}\dwarr & \avestd{0.35}{0.02} & \avestd{~~0.32}{0.01}\dwarr \\
        \quad Cosine between pos vs. neg     & \avestd{0.48}{0.03} & \avestd{~~0.35}{0.03}\dwarr & \avestd{0.55}{0.01} & \avestd{~~0.35}{0.02}\dwarr & \avestd{0.39}{0.01} & \avestd{~~0.37}{0.02}\dwarr \\
        \midrule
        Pre-logit weight/templates $\ma{W}$ \\
        \quad $\ell_2$-norm                  & \avestd{1.93}{0.28} & \avestd{~~1.57}{0.03}\dwarr & \avestd{1.84}{0.08} & \avestd{~~1.39}{0.02}\dwarr & \avestd{1.71}{0.04} & \avestd{~~1.39}{0.01}\dwarr \\
        \quad Abs cosine                     & \avestd{0.17}{0.12} & \avestd{~~0.10}{0.06}\dwarr & \avestd{0.31}{0.11} & \avestd{~~0.10}{0.10}\dwarr & \avestd{0.23}{0.08} & \avestd{~~0.05}{0.03}\dwarr \\
        \quad Abs cosine between pos vs. pos & \avestd{0.13}{0.10} & \avestd{~~0.10}{0.07}\dwarr & \avestd{0.27}{0.11} & \avestd{~~0.10}{0.11}\dwarr & \avestd{0.17}{0.04} & \avestd{~~0.05}{0.04}\dwarr \\
        \quad Abs cosine between pos vs. neg & \avestd{0.31}{0.06} & \avestd{~~0.10}{0.05}\dwarr & \avestd{0.40}{0.06} & \avestd{~~0.09}{0.04}\dwarr & \avestd{0.32}{0.04} & \avestd{~~0.05}{0.03}\dwarr \\
        \bottomrule
    \end{tabular}
    \caption{$\ell_2$-norm and cosine similarity of pre-logit representations and templates. 
    ``pos'' means the positive types (i.e., entity types); ``neg'' means the negative type (i.e., non-entity type). 
    \uparr/\dwarr\ indicates that \ours\ presents a metric higher/lower than its shallow counterpart. 
    All the metrics are first averaged within each experiment, and then averaged over five independent experiments, reported with subscript standard deviations.}
    \label{tab:l2-and-cos}
\end{table*}

\begin{figure*}[!t]
    \centering
    \begin{subfigure}{0.32\textwidth}
    \centering
    \includegraphics[width=\textwidth]{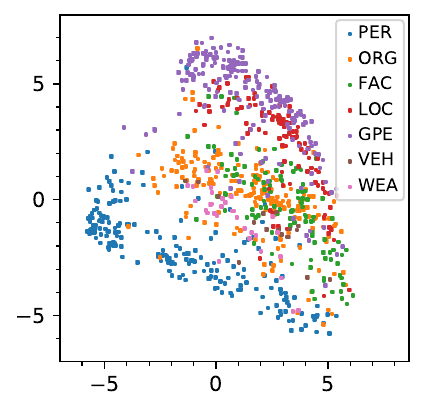}
    \caption{ACE 2004, Shallow}
    \end{subfigure}
    \begin{subfigure}{0.32\textwidth}
    \centering
    \includegraphics[width=\textwidth]{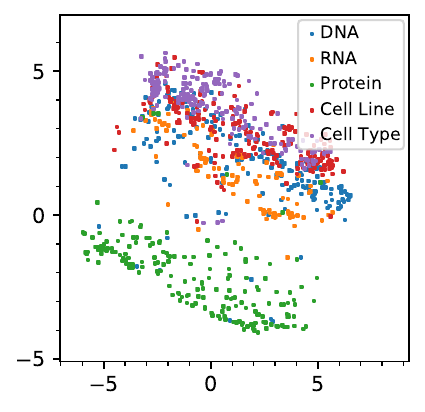}
    \caption{GENIA, Shallow}
    \end{subfigure}
    \begin{subfigure}{0.32\textwidth}
    \centering
    \includegraphics[width=\textwidth]{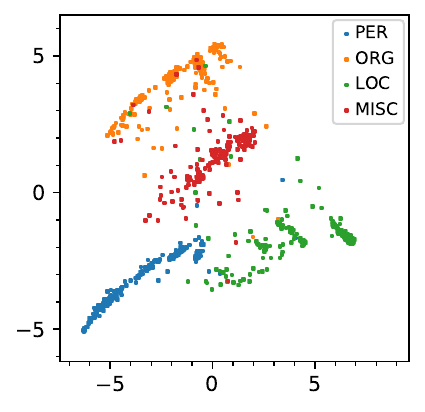}
    \caption{CoNLL 2003, Shallow}
    \end{subfigure}
    \centering
    \begin{subfigure}{0.32\textwidth}
    \centering
    \includegraphics[width=\textwidth]{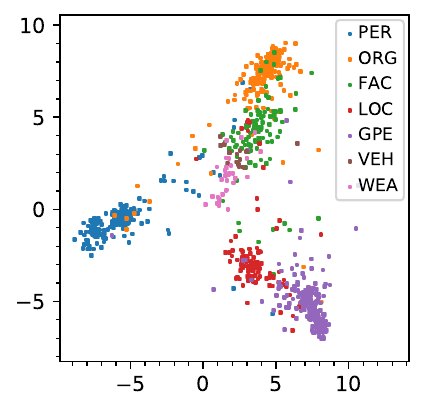}
    \caption{ACE 2004, Deep}
    \end{subfigure}
    \begin{subfigure}{0.32\textwidth}
    \centering
    \includegraphics[width=\textwidth]{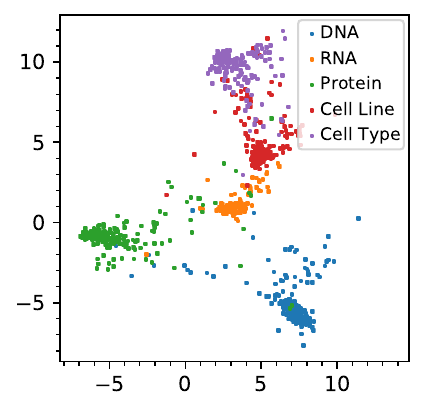}
    \caption{GENIA, Deep}
    \end{subfigure}
    \begin{subfigure}{0.32\textwidth}
    \centering
    \includegraphics[width=\textwidth]{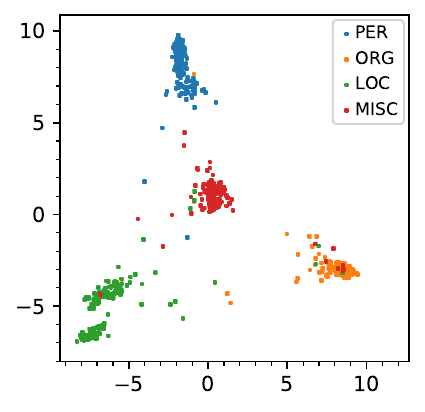}
    \caption{CoNLL 2003, Deep}
    \end{subfigure}
    \caption{PCA visualization of pre-logit span representations of entities in the testing set.}
    \label{fig:pca}
\end{figure*}

\subsection{Visualization}
Figure~\ref{fig:pca} visualizes the span representations dimension-reduced by principal component analysis (PCA). The results are quite impressive. The representations by shallow aggregation are scattered over the plane. Although they are largely clustered by categories, the boundaries are mixed and ambiguous. In contrast, the deep span representations group by relatively clear and tight clusters, corresponding to the ground-truth categories. Except for some outliers, each pair of the clusters can be easily separable in this projected plane. This is also consistent to the aforementioned finding that deep span representations have high within-class similarities but low between-class similarities. As a linear dimensionality reduction, PCA results indicate whether and how the features are linearly separable. Note that the pre-logit representations are the last ones before the logits, so the linear separability is crucial to the classification performance.

\section{Discussion and Conclusion}
Neural NLP models have been rigidly adhere to the paradigm where an encoder produces token-level representations, and a task-specific decoder receives these representations, computes the loss and yields the outputs~\citep{collobert2011natural}. This design works well on most, if not all, NLP tasks; and it may also deserve a credit for facilitating NLP pretraining~\citep{peters-etal-2018-deep, devlin-etal-2019-bert}, since such common structure in advance bridges the pretraining and downstream phases. 

However, this paradigm may be sub-optimal for specific tasks. In span-based NER (or information extraction from a broader perspective), the smallest modeling unit should be spans instead of tokens; and thus the span representations should be crucial. This originally motivates our \ours. In addition, \ours\ also successfully shows how to exploit the pretrained weights beyond the original Transformer structure, i.e., adapting the weights from computing token representations for span representations. We believe that adding span representation learning in the pretraining stage will further contribute positively. 

In conclusion, deep and span-specific representations can significantly boost span-based neural NER models. Our \ours\ achieves SOTA results on six well-known NER benchmarks; the model presents pronounced effect on long-span entities and nested structures. Further analysis shows that the resulting deep span representations are well structured and easily separable in the feature space.

\section{Limitations}
To some extent, \ours\ pursues performance and interpretability over computational efficiency. The major computational cost of a Transformer encoder is on the multihead attention module and FFN. For a $T$-length input and a $d$-dimensional Transformer encoder, the per-layer complexities of the multihead attention and FFN are of order $O (T^2 d)$ and $O (T d^2)$, respectively. When the maximum span size $K \ll T$, our span Transformer brings additional $O (K^2 T d)$ complexity on the attention module, and $O (K T d^2)$ complexity on the FFN. Empirically, training a \ours\ consumes about five times the time for a shallow model of a same scale. However, this issue can be mitigated if we use fewer layers for the span Transformer (Subsection~\ref{subsec:ablation}). 

As noted, we empirically choose the maximum span size $K$ such that it covers most entities in the training and development splits. From the perspective of $F_1$ score, this heuristic works well, and \ours\ performs favourably on long-span entities as long as they are covered. However, the entities with extreme lengths beyond $K$ will be theoretically irretrievable.

\section*{Acknowledgements}
This work is supported by Zhejiang Provincial Natural Science Foundation of China (No. LQ23F020005), National Natural Science Foundation of China (No. 62106248). We thank the anonymous reviewers for their insightful and constructive comments.

\bibliography{anthology,custom,references}
\bibliographystyle{acl_natbib}

\newpage
\appendix
\section{(Shallowly) Aggregating Functions} \label{app:aggregating}
Given token representations $\ma{H} \in \mathbb{R}^{T \times d}$, a model can shallowly aggregate them in corresponding positions to construct span representations. Formally, the span representation of $(i, j)$ can be built by: 

\paragraph{Max-pooling.} Applying max-pooling to $\ma{H}_{[i:j]}$ over the first dimension. 

\paragraph{Mean-pooling.} Applying mean-pooling to $\ma{H}_{[i:j]}$ over the first dimension. 

\paragraph{Multiplicative Attention.} Computing 
\begin{equation*}
\softmax \left( \ve{u}^\trans \tanh \left( \ma{W} \ma{H}_{[i:j]}^\trans \right) \right) \ma{H}_{[i:j]}, 
\end{equation*}
where $\ma{W}$ and $\ve{u}$ are learnable parameters. 

\paragraph{Additive Attention.} Computing
\begin{equation*}
\softmax \left( \ve{u}^\trans \tanh \left( \ma{W} \left( \ma{H}_{[i:j]} \oplus \ve{v} \right)^\trans \right) \right) \ma{H}_{[i:j]}, 
\end{equation*}
where $\ma{W}$, $\ve{u}$ and $\ve{v}$ are learnable parameters; $\oplus$ means concatenation over the second dimension, and vector $\ve{v}$ should be repeated for $j-i$ times before the concatenation. 

In general, either multiplicative or additive attention computes normalized weights over the sequence dimension of span $(i, j)$, where the weights are dependent on the values of $\ma{H}_{[i:j]}$; and then applies the weights to $\ma{H}_{[i:j]}$, resulting in weighted average values.

\section{Datasets} \label{app:datasets}

\begin{table*}[t]
    \centering \small
    \begin{tabular}{llrrrrrr}
        \toprule
         &  & ACE04 & ACE05 & GENIA & KBP17 & CoNLL03 & OntoNotes 5 \\% & Weibo & Resume \\
        \midrule
        \multirow{4}{*}{\#Sent.} & All   & 8,507   & 9,341   & 18,546  & 15,358  & 22,137  & 76,714    \\% & 1,890   & 4,761 \\
                                 & Train & 6,799   & 7,336   & 15,023  & 10,546  & 14,987  & 59,924    \\% & 1,350   & 3,821 \\
                                 & Dev.  & 829     & 958     & 1,669   & 545     & 3,466   & 8,528     \\% & 270     & 463 \\
                                 & Test  & 879     & 1,047   & 1,854   & 4,267   & 3,684   & 8,262     \\% & 270     & 477 \\
        \midrule
        \#Type                   &       & 7       & 7       & 5       & 5       & 4       & 18        \\% & 8       & 8 \\
        \#Token                  &       & 173,796 & 176,575 & 471,264 & 300,345 & 302,811 & 1,388,955 \\% & 103,129 & 153,089 \\
        \#Entity                 &       & 27,749  & 30,931  & 56,046  & 45,714  & 35,089  & 104,151   \\% & 2,702   & 16,567 \\
        \#Nested                 &       & 7,832   & 7,460   & 5,431   & 7,479   & --      & --        \\% & --      & -- \\
        \bottomrule
    \end{tabular}
    \caption{Descriptive statistics of datasets. \#Sent. denotes the number of sentences; \#Type denotes the number of entity types; \#Token denotes the number of tokens; \#Entity denotes the number of entities; \#Nested denotes the number of entities that are nested in other entities.}
    \label{tab:descriptive-statstics}
\end{table*}

\begin{table*}[t]
    \centering \small
    \begin{tabular}{lcccccc}
        \toprule
         & ACE04 & ACE05 & GENIA & KBP17 & CoNLL03 & OntoNotes 5 \\% & Weibo & Resume \\
        \midrule
        PLM                              & \multicolumn{6}{c}{RoBERTa-base} \\% & \multicolumn{2}{c}{MacBERT-base} \\
        Maximum span size                & 25   & 25   & 18   & 16   & 10   & 16   \\% & 15   & 25 \\
        Initial aggregation              & Max  & Max  & Max  & Max  & Max  & Max  \\% & Max  & Max \\
        LSTM hidden size                 & 400  & 400  & 400  & 400  & 400  & 400  \\% & 400  & 400 \\
        LSTM layers                      & 1    & 1    & 1    & 1    & 1    & 1    \\% & 1    & 1 \\
        FFN hidden size                  & 300  & 300  & 300  & 300  & 300  & 300  \\% & 300  & 300 \\
        FFN layers                       & 1    & 1    & 1    & 1    & 1    & 1    \\% & 1    & 1 \\
        Boundary smoothing $\varepsilon$ & 0.1  & 0.1  & 0.1  & 0.1  & 0.1  & 0.1  \\% & 0.2  & 0.2 \\
        Number of epochs                 & 50   & 50   & 30   & 50   & 50   & 30   \\% & 50   & 50 \\
        Learning rate \\
        \quad Pretrained weights         & 2e-5 & 2e-5 & 4e-5 & 2e-5 & 2e-5 & 2e-5 \\% & 2e-5 & 2e-5 \\
        \quad Other weights              & 2e-3 & 2e-3 & 2e-3 & 2e-3 & 2e-3 & 2e-3 \\% & 2e-3 & 2e-3 \\
        Batch size                       & 48   & 48   & 16   & 48   & 48   & 16   \\% & 48   & 48 \\
        \midrule
        Number of parameters (M)         & \multicolumn{6}{c}{212.9 (w/o weight sharing); or 126.3 (w/ weight sharing)} \\
        Training time (hours on A6000)   & 13.9 & 15.0 & 8.3  & 16.2 & 3.3  & 15.3 \\
        \bottomrule
    \end{tabular}
    \caption{Hyperparameters and computational costs of main experiments.}
    \label{tab:hyperparameters}
\end{table*}

\paragraph{ACE 2004 and ACE 2005} are two English nested NER datasets, either of which contains seven entity types, i.e., Person, Organization, Facility, Location, Geo-political Entity, Vehicle, Weapon. Our data processing and splits follow \citet{lu-roth-2015-joint}. 

\paragraph{GENIA}~\citep{kim2003genia} is a nested NER corpus on English biological articles. Our data processing follows \citet{lu-roth-2015-joint}, resulting in five entity types (DNA, RNA, Protein, Cell line, Cell type); data splits follow \citet{yan-etal-2021-unified-generative} and \citet{li2022unified}. 

\paragraph{KBP 2017}~\citep{ji2017overview} is an English nested NER corpus including text from news, discussion forum, web blog, tweets and scientific literature. It contains five entity categories, i.e., Person, Geo-political Entity, Organization, Location, and Facility. Our data processing and splits follow \citet{lin-etal-2019-sequence} and \citet{shen-etal-2022-parallel}. 

\paragraph{CoNLL 2003}~\citep{tjong-kim-sang-de-meulder-2003-introduction} is an English flat NER benchmark with four entity types, i.e., Person, Organization, Location and Miscellaneous. We use the original data splits for experiments. 

\paragraph{OntoNotes 5} is a large-scale English flat NER benchmark, which has 18 entity types. Our data processing and splits follow \citet{pradhan-etal-2013-towards}. 

Table~\ref{tab:descriptive-statstics} presents the descriptive statistics of the datasets.

\section{Implementation Details} \label{app:implementation}
\paragraph{Hyperparameters.} We choose RoBERTa~\citep{liu2019roberta} as the PLM to initialize the weights in the Transformer blocks and span Transformer blocks. The PLMs used in our main experiments are all of the \texttt{base} size (768 hidden size, 12 layers). 

For span Transformer blocks, the maximum span size $K$ is specifically determined for each dataset. In general, a larger $K$ would improve the recall performance (entities longer than $K$ will never be recalled), but significantly increase the computation cost. We empirically choose $K$ such that it covers most entities in the training and development splits. For example, most entities are short in CoNLL 2003, so we use $K = 10$; while entities are relatively long in ACE 2004 and ACE 2005, so we use $K = 25$. We use max-pooling as the initial aggregating function. 

We find it beneficial to additionally include a BiLSTM~\citep{hochreiter1997long} before passing the span representations to the entity classifier. The BiLSTM has one layer and 400 hidden states, with dropout rate of 0.5. In the entity classifier, the FFN has one layer and 300 hidden states, with dropout rate of 0.4, and the activation is ReLU~\citep{krizhevsky2012imagenet}. In addition, boundary smoothing~\citep{zhu-li-2022-boundary} with $\varepsilon = 0.1$ is applied to loss computation. 

We train the models by the AdamW optimizer~\citep{loshchilov2018decoupled} for 50 epochs with the batch size of 48. Gradients are clipped at $\ell_2$-norm of 5~\citep{pascanu2013difficulty}. The learning rates are 2e-5 and 2e-3 for pretrained weights and randomly initialized weights, respectively; a scheduler of linear warmup is applied in the first 20\% steps followed by linear decay. For some datasets, a few hyperparameters are further tuned and thus slightly different from the above ones. 

The experiments are run on NVIDIA RTX A6000 GPUs. More details on the hyperparameters and computational costs are reported in Table~\ref{tab:hyperparameters}.

\paragraph{Evaluation.} An entity is considered correctly recognized if its predicted type and boundaries exactly match the ground truth. 

The model checkpoint with the best $F_1$ score throughout the training process on the development split is used for evaluation. The evaluation metrics are micro precision rate, recall rate and $F_1$ score on the testing split. Unless otherwise noted, we run each experiment for five times and report the average metrics with corresponding standard deviations.

\section{Categorical Results} \label{app:categorical}
Table~\ref{tab:categorical} lists the category-specific results on ACE 2004, GENIA and CoNLL 2003. As a strong baseline, the classic biaffine model~\citep{yu-etal-2020-named} is re-implemented with PLM and hyperparameters consistent with our \ours; note that our re-implementation achieves higher performances than the scores reported in the original paper. The categorical results show that \ours\ outperforms the biaffine model across almost all the categories of the three datasets, except for Geo-political Entity and Vehicle from ACE 2004. 

\begin{table}[t]
    \centering \small
    \begin{tabular}{cccc}
        \toprule
         & \multicolumn{3}{c}{ACE 2004} \\
        \cmidrule(lr){2-4} 
         & Biaffine & \ours & $\Delta$ \\
        \midrule
        \texttt{PER} & \avestd{91.35}{0.10} & \avestd{91.50}{0.16} & \textcolor{casgreen}{+0.14} \\
        \texttt{ORG} & \avestd{82.49}{0.54} & \avestd{83.59}{0.33} & \textcolor{casgreen}{+1.10} \\
        \texttt{FAC} & \avestd{71.18}{1.30} & \avestd{72.28}{1.43} & \textcolor{casgreen}{+1.10} \\
        \texttt{LOC} & \avestd{73.38}{1.55} & \avestd{75.22}{0.86} & \textcolor{casgreen}{+1.84} \\
        \texttt{GPE} & \avestd{89.30}{0.36} & \avestd{89.16}{0.44} & \textcolor{casred}{-0.15} \\
        \texttt{VEH} & \avestd{87.18}{2.66} & \avestd{84.35}{3.58} & \textcolor{casred}{-2.83} \\
        \texttt{WEA} & \avestd{70.87}{4.18} & \avestd{79.88}{1.41} & \textcolor{casgreen}{+9.00} \\
        \midrule
        Overall      & \avestd{87.64}{0.19} & \avestd{\textbf{88.05}}{0.18} & \textcolor{casgreen}{+0.41} \\
        \bottomrule 
        \toprule
         & \multicolumn{3}{c}{GENIA} \\
        \cmidrule(lr){2-4} 
         & Biaffine & \ours & $\Delta$ \\
        \midrule
        DNA       & \avestd{77.50}{0.44} & \avestd{77.75}{0.56} & \textcolor{casgreen}{+0.26} \\
        RNA       & \avestd{83.75}{0.68} & \avestd{85.02}{1.31} & \textcolor{casgreen}{+1.27} \\
        Protein   & \avestd{84.00}{0.36} & \avestd{84.57}{0.26} & \textcolor{casgreen}{+0.57} \\
        Cell Line & \avestd{74.60}{0.48} & \avestd{76.39}{0.96} & \textcolor{casgreen}{+1.79} \\
        Cell Type & \avestd{75.89}{0.46} & \avestd{76.10}{0.48} & \textcolor{casgreen}{+0.22} \\
        \midrule
        Overall   & \avestd{80.93}{0.24} & \avestd{\textbf{81.46}}{0.25} & \textcolor{casgreen}{+0.53} \\
        \bottomrule 
        \toprule
         & \multicolumn{3}{c}{CoNLL 2003} \\
        \cmidrule(lr){2-4} 
         & Biaffine & \ours & $\Delta$ \\
        \midrule
        \texttt{PER}  & \avestd{96.74}{0.23} & \avestd{96.94}{0.11} & \textcolor{casgreen}{+0.20} \\
        \texttt{ORG}  & \avestd{92.92}{0.13} & \avestd{93.09}{0.14} & \textcolor{casgreen}{+0.17} \\
        \texttt{LOC}  & \avestd{94.83}{0.18} & \avestd{94.96}{0.13} & \textcolor{casgreen}{+0.12} \\
        \texttt{MISC} & \avestd{83.83}{0.28} & \avestd{84.52}{0.56} & \textcolor{casgreen}{+0.69} \\
        \midrule
        Overall       & \avestd{93.37}{0.10} & \avestd{\textbf{93.64}}{0.06} & \textcolor{casgreen}{+0.27} \\
        \bottomrule 
    \end{tabular}
    \caption{Categorical $F_1$ scores by biaffine model and \ours. All the results are average scores of five independent runs, with subscript standard deviations.}
    \label{tab:categorical}
\end{table}

\section{Results on Chinese NER} \label{app:chinese-flat-res}
Table~\ref{tab:chinese-flat-res} shows the experimental results on two Chinese flat NER datasets: Weibo NER~\citep{peng-dredze-2015-named} and Resume NER~\citep{zhang-yang-2018-chinese}. \ours\ achieves 72.64\% and 96.72\% best $F_1$ scores on the two benchmarks; they are also quite close to the recently reported SOTA results. 

\begin{table}[t]
    \centering \small
    \begin{tabular}{lccl}
        \toprule
        \multicolumn{4}{c}{Weibo NER} \\
        \midrule
        Model & Prec. & Rec. & ~~F1 \\
        \midrule
        \citet{ma-etal-2020-simplify}       & -- & -- & 70.50 \\
        \citet{li-etal-2020-flat}           & -- & -- & 68.55 \\
        \citet{wu-etal-2021-mect}           & -- & -- & 70.43 \\
        \citet{li2022unified}\textdaggerdbl & 70.84 & 73.87 & 72.32 \\
        \citet{zhu-li-2022-boundary}        & 70.16 & 75.36 & \textbf{72.66} \\
        \midrule
        \ours \textdagger                   & 74.56 & 70.81 & 72.64 \\
        \ours \textdaggerdbl                & 71.09 & 71.58 & \avestd{71.30}{0.86} \\
        \bottomrule
        \toprule
        \multicolumn{4}{c}{Resume NER} \\
        \midrule
        Model & Prec. & Rec. & ~~F1 \\
        \midrule
        \citet{ma-etal-2020-simplify}       & 96.08 & 96.13 & 96.11 \\
        \citet{li-etal-2020-flat}           & -- & -- & 95.86 \\
        \citet{wu-etal-2021-mect}           & -- & -- & 95.98 \\
        \citet{li2022unified}\textdaggerdbl & 96.96 & 96.35 & 96.65 \\
        \citet{zhu-li-2022-boundary}        & 96.63 & 96.69 & 96.66 \\
        \midrule
        \ours \textdagger                   & 96.69 & 96.75 & \textbf{96.72} \\
        \ours \textdaggerdbl                & 96.44 & 96.58 & \avestd{96.51}{0.17} \\
        \bottomrule
    \end{tabular}
    \caption{Results of Chinese flat entity recognition. \textdagger~means the best score; \textdaggerdbl~means the average score of multiple independent runs; the subscript number is the corresponding standard deviation.}
    \label{tab:chinese-flat-res}
\end{table}

\section{Additional Ablation Studies} \label{app:ablation}

\paragraph{Weight Sharing.} As described, the span Transformer shares the same structure with the Transformer, but their weights are independent and separately initialized from the PLM. A straightforward idea is to tie the corresponding weights between these two modules, which can reduce the model parameters and conceptually performs as a regularization technique. 

As reported in Table~\ref{tab:weight-sharing}, sharing the weights results in higher $F_1$ scores on ACE 2004, but lower scores on GENIA and CoNLL 2003; the performance differences are largely insignificant. Note that with weight sharing, the span Transformer fully reuses the weights of the standard Transformer, thus requires no additional parameters. This suggests that the performance improvements of \ours\ are from better span representations instead of increased model size. 

In addition to the performance, weight sharing has very limited effect on reducing training and inference cost --- both the forward and backward computations remain almost unchanged. However, it does effectively halve the parameter number; hence, weight sharing would be a preferred option when the storage space is limited. 

\begin{table}[t]
    \centering \small
    \begin{tabular}{cccc}
        \toprule
        Weight Sharing & ACE04 & GENIA & CoNLL03 \\
        \midrule
        \underline{\texttimes} & \avestd{88.05}{0.18} & \avestd{\textbf{81.46}}{0.25} & \avestd{\textbf{93.64}}{0.06} \\
        \checkmark & \avestd{\textbf{88.11}}{0.23} & \avestd{81.19}{0.26} & \avestd{93.43}{0.17} \\
        \bottomrule
    \end{tabular}
    \caption{The effect of weight sharing. The underlined specification is the one used in our main experiments. All the results are average scores of five independent runs, with subscript standard deviations.}
    \label{tab:weight-sharing}
\end{table}

\paragraph{Initial Aggregation Functions.} We test different functions for initial aggregation in the span Transformer. As shown in Table~\ref{tab:ablation}, max-pooling outperforms all other alternatives, although the performance differences are limited. This result is different from that in the shallow setting, where max-pooling underperforms multiplicative and additive attentions (Table~\ref{tab:depth}). 

One possible explanation is that, the span Transformer blocks with weights initialized from the PLM have already performed very sophisticated multi-head attentions through multiple layers, so one extra attention layer with randomly initialized weights cannot take positive effect further, or even plagues the model. Max-pooling, on the other hand, is parameter-free and thus more advantageous in this case. 

\paragraph{Pretrained Language Models.} Table~\ref{tab:ablation} also lists the results by alternative PLMs. In general, BERT-base and BERT-large~\citep{devlin-etal-2019-bert} underperforms RoBERTa-base~\citep{liu2019roberta}, while RoBERTa-large can further improve the performance by 0.54, 0.52 and 0.06 percentage $F_1$ scores on ACE 2004, GENIA and CoNLL 2003, respectively. These results are consistent with \citet{zhu-li-2022-boundary}, confirming the superiority of RoBERTa in NER tasks. 

Since GENIA is a biological corpus, some previous studies use BioBERT on this benchmark~\citep{shen-etal-2021-locate,shen-etal-2022-parallel}. We also test BioBERT with \ours\ on GENIA. The results show that BioBERT can achieve performance competitive to RoBERTa. 

\paragraph{BiLSTM and Boundary Smoothing.} As presented in Table~\ref{tab:ablation}, removing the BiLSTM layer will result in a drop of 0.2--0.4 percentage $F_1$ scores. In addition, replacing boundary smoothing~\citep{zhu-li-2022-boundary} with the standard cross entropy loss will reduce the $F_1$ scores by similar magnitudes. 

\begin{table}[t]
    \centering \small
    \begin{tabular}{lccc}
        \toprule
         & ACE04 & GENIA & CoNLL03 \\
        \midrule
        Initial agg. \\
        \quad \underline{w/ max-pooling} & \avestd{88.05}{0.18} & \avestd{81.46}{0.25} & \avestd{93.64}{0.06} \\
        \quad w/ mean-pooling   & \avestd{87.77}{0.31} & \avestd{81.31}{0.14} & \avestd{93.58}{0.11} \\
        \quad w/ mul. attention & \avestd{87.67}{0.29} & \avestd{81.36}{0.17} & \avestd{93.55}{0.09} \\
        \quad w/ add. attention & \avestd{87.90}{0.11} & \avestd{81.20}{0.12} & \avestd{93.56}{0.12} \\
        \midrule
        PLM \\
        \quad \underline{w/ RoBERTa-b} & \avestd{88.05}{0.18} & \avestd{81.46}{0.25} & \avestd{93.64}{0.06} \\
        \quad w/ RoBERTa-l & \avestd{\textbf{88.59}}{0.27} & \avestd{\textbf{81.98}}{0.41} & \avestd{\textbf{93.70}}{0.12} \\
        \quad w/ BERT-b    & \avestd{86.38}{0.20} & \avestd{79.13}{0.16} & \avestd{91.90}{0.08} \\
        \quad w/ BERT-l    & \avestd{87.73}{0.30} & \avestd{79.27}{0.20} & \avestd{92.79}{0.14} \\
        \quad w/ BioBERT-b &  & \avestd{81.52}{0.26} & \\
        \quad w/ BioBERT-l &  & \avestd{81.78}{0.26} & \\
        \midrule
        Others \\
        \quad w/o BiLSTM   & \avestd{87.71}{0.05} & \avestd{81.29}{0.33} & \avestd{93.42}{0.11} \\
        % \quad w/o FFN      &  &  & \\
        \quad w/o BS       & \avestd{87.66}{0.26} & \avestd{81.02}{0.27} & \avestd{93.45}{0.17} \\
        \bottomrule
    \end{tabular}
    \caption{Results of ablation studies. ``b'' and ``l'' mean the PLM sizes of \texttt{base} and \texttt{large}, respectively; for \texttt{large} PLM, span Transformer has 12 layers. ``BS'' means boundary smoothing. The underlined specification is the one used in our main experiments. All the results are average scores of five independent runs, with subscript standard deviations.}
    \label{tab:ablation}
\end{table}

\end{document}